\documentclass[journal]{IEEEtran}

\usepackage[utf8]{inputenc}
\usepackage[numbers,sort&compress]{natbib}

\usepackage[table]{xcolor}
\usepackage{times}
\usepackage{epigraph}
\usepackage{epsfig}
\usepackage{graphicx}
\usepackage{rotating}
\usepackage{amsmath}
\usepackage{amssymb}
\usepackage{booktabs}
\usepackage{float}
\usepackage{placeins}
\usepackage{mathabx}
\usepackage{multirow}
\usepackage{balance}
\usepackage{multirow}
\usepackage{soul}
\usepackage{subfigure}
\usepackage[flushleft]{threeparttable}
\usepackage{textcomp}
\usepackage{enumitem}
\usepackage{url}

\setlist[enumerate]{leftmargin=*}
\setlist[itemize]{leftmargin=*}
\setlist[description]{leftmargin=*}

\begin{document}

\title{Deep-Learning Ensembles for Skin-Lesion Segmentation, Analysis, Classification: \\ RECOD Titans at ISIC Challenge 2018}

\author{
    Alceu Bissoto$^\ast$\thanks{$^\ast$A. Bissoto is the main author of Part 2; F. Perez is the main author of Part 3; V. Ribeiro is the main author of Part 1.}, Fábio Perez$^\ast$, Vinícius Ribeiro$^\ast$, Michel Fornaciali, Sandra Avila, Eduardo Valle$^\dag$
    \vspace{-0.75cm}

    \thanks{      
        F. Perez, V. Ribeiro, M. Fornaciali, and E. Valle are affiliated the Department of Computer Engineering and Industrial Automation (DCA) of the School of Electrical and Computing Engineering (FEEC), University of Campinas (UNICAMP), Brazil.       
        A. Bissoto and S. Avila are affiliated to the Institute of Computing (IC), UNICAMP.     
        All authors are also affiliated to the RECOD Lab (recodbr.wordpress.com).
        
        $^\dag$Contact author: dovalle@dca.fee.unicamp.br, mail@eduardovalle.com.
    }
}

%\markboth{Journal of \LaTeX\ Class Files,~Vol.~11, No.~4, December~2012}%
%{Shell \MakeLowercase{\textit{et al.}}: Bare Demo of IEEEtran.cls for Journals}

\maketitle

% \begin{IEEEkeywords}
%     Melanoma Screening, Dermoscopy, Deep Learning, Transfer Learning.
% \end{IEEEkeywords}

\section{History}

Our team has worked on skin lesion analysis since early 2014~\cite{fornaciali2014statistical}, and has employed deep learning with transfer learning for that task since 2015~\cite{carvalho2015}. From 2016 onwards, the community moved from traditional techniques towards deep learning, following the general trend of computer vision~\cite{fornaciali2016towards}. Deep learning poses a challenge for medical applications, as it needs very large training sets. Thus, transfer learning becomes crucial for success in those applications, motivating our paper for ISBI 2017~\cite{menegola2017knowledge}. Until 2017, the contribution of each factor of a deep learning solution (e.g., model choice, dataset size, data augmentation, image normalization, etc.) to the performance of a skin lesion classifier was not evident. We cleared such question by extensively analyzing several combinations of architectures, dataset sizes, and other eight relevant aspects~\cite{valle2017data}.
  
We participated in the ISIC Challenge 2017, being ranked in 1st place for melanoma classification and 5th place for skin lesion segmentation~\cite{menegola2017recod}. In 2018, for the first time, we participated in all three tasks.Although  our team has a long experience with skin-lesion classification (Task~3) and moderate experience with lesion segmentation (Task~1), this Challenge was the very first time we worked on attribute detection (Task~2).

\section{Generalities}

\subsection{Strategy}

We aimed, from the start, at deep learning solutions for all tasks. We know from experience that the success factors for a competitive deep learning approach are data availability and model depth~\cite{menegola2017recod,valle2017data}. To improve our chances, we also introduced two original contributions --- synthetic lesions generation and stronger data augmentation approaches --- to boost the models training. Such contributions will be detailed next. Participating in the Challenge brings our sportive desire to squeeze the models for their best performance --- as always, we temper that goal with aesthetic considerations, avoiding as much as possible kludges and added complexity. Added complexity has to bring proportional improvements over the metrics, or we will prefer the simpler model.

Each task allowed up to 3 distinct submissions. We used them to contrast models trained with extra data with models trained with challenge-data only, or to compare different ways to ensemble the final solutions.

\subsection{Data}

In previous work, we showed that the training set size responds by almost 50\% of the variation on the prediction power of the classifier~\cite{valle2017data}. The freedom to use external sources enabled us to gather more data to boost our models. First, we restricted ourselves to publicly available (for free, or for a fee) sources with high-quality images: 

\begin{description}
  \item [ISIC 2018 Challenge~{\normalfont\cite{codella2018skin,Tschandl2018_HAM10000}}] the official challenge dataset, with 10,015 dermoscopic images.
  \item [ISIC Archive\footnote{The ISIC Archive: http://isdis.net/isic-project/}] with over 13,000 dermoscopic images.
  \item [Interactive Atlas of Dermoscopy~{\normalfont\cite{argenziano2002dermoscopy}}] with 1,000+ clinical cases, each with dermoscopic, and close-up clinical images.           
  \item [Dermofit Image Library~{\normalfont\cite{ballerini2013color}}] with 1,300 images.
  \item [PH2 Dataset~{\normalfont\cite{mendoncca2013ph}}] with 200 dermoscopic images.       
\end{description}

However, due to the extreme imbalance of the dataset, we decided to gather extra images for the severely underrepresented classes (namely Actinic keratosis, Basal cell carcinoma, Dermatofibroma, and Vascular lesion). We found images browsing sources on the web, and asking for contributions from partner researchers in Medical Science (acknowledged in the final section). The web sources were \textbf{Dermatology Atlas} (www.atlasdermatologico.com.br), \textbf{Derm101} (www.derm101.com), \textbf{DermIS} (www.dermis.net/dermisroot). With that extra effort, we acquired additional 631 images, being 414 BCC, 26 AKIEC, 132 DF and 59 VASC. The final dataset continued seriously unbalanced, but the proportion of underrepresented classes grew considerably. Our final dataset had 30,726 images. 

We evaluated our extra data (\textbf{full}) on Task 1 (with 18,179 images, those with segmentation ground-truth), and Task 3 (with 30,324 images, those with diagnosis label). We did not have ground truth for Task 2 --- other than the 2017 Challenge data, which we briefly considered employing --- so for this task, we did not use extra data. For the three tasks we also made submissions using only Challenge data (\textbf{only}). 

After picking a dataset, we divided it the into 3 splits, for each task: 10\% for holdout (for our internal model selection) and the remaining 90\% for training. The training split was further divided into five 10\%-validation/90\%-training different splits (at random, not using cross-validation folds). We considered case numbers, aliases, and near-duplicates in the split division, to minimize contaminations across splits.

We used the holdout sets to select the models. We used the metrics observed in the holdout sets to identify strong release candidate models and/or good bets for a meta-learning phase. Although the official validation data was very limited on this year's Challenge, we still used its scores as ancillary estimates.

The exact datasets and splits, for each task, will be listed, image by image, in our \textbf{code repository}\footnote{https://github.com/learningtitans/isic2018-\{part1,part2,part3\} (available soon)}.

\subsection{Experimental \st{Design} Tactics}
         
Our starting point was our last work on how to design powerful deep-learning classifiers for skin lesions~\cite{valle2017data}. We evaluated the main factors that vary on the approaches found in literature: use of transfer learning, model architecture, train dataset, image resolution, type of data augmentation, input normalization, use of segmentation, duration of training, additional use of SVM, and test data augmentation.
    
For the challenge, there was no time to perform such a detailed study --- involving significance tests over a full-factorial design --- but we wanted to make sound decisions along the way. We decided to use our previous study to eliminate many choices and perform much-reduced designs, involving less than a handful of factors. We will describe the factors (and their levels) in the task sections.

The team used the Slack collaboration tool as the main channel for communication. We coordinated the tasks with Google Docs and shared the results of each intermediate experiment with Google Sheets. We used code version control (with git) to facilitate future reproduction of intermediate steps.

\subsection{Notable Novelties}

The models we proposed this year have several technical advances in comparison with the models we submitted last year: deeper architectures, changes in frameworks, better training craftsmanship, etc. In this section, however, we showcase the most exciting scientific novelties.

For this year we took advantage of our recent results regarding new approaches for data augmentation: (a) image processing of real skin lesion images~\cite{perez2018data}, and (b) synthetic skin lesions using GANs (Generative Adversarial Networks)~\cite{bissoto2018skin}. 

In work (a), we investigated the impact of 13 image processing-based scenarios of data augmentation for melanoma classification. Scenarios include traditional color and geometric transforms, and more unusual augmentations such as elastic transforms, random erasing and a novel augmentation that mix two different lesions. Using our participation on ISIC Challenge 2017 (with Inception-v4) with as baseline, we observed similar performance using the new data augmentation methods, but without using external data. That is, the image processing data augmentation methods were equal to the performance of the model trained with external data (which we know that has a huge impact on the classifier prediction power~\cite{valle2017data}). Among all experiments and scenarios, scenario J (please refer to~\cite{perez2018data} for details) leads to better performance and was the one introduced in the experiments of the competition (only in Task 3).

In work (b), we created fake high-resolution (1024$\times$512 pixels) skin lesion samples, aiming to extend the training set artificially. To do that, we used GANs to teach the network the malignancy markers and also incorporating the specificities of a lesion border. We inputted such information directly to the network, using a semantic map and an instance map. Semantic maps are blobs that show the presence and the location of the five malignancy markers within the same lesions' segmentation masks. Instance maps take information from superpixels, which group similar pixels creating visually meaningful blobs, limiting each unit regarding their meaning. Please refer to~\cite{bissoto2018skin} for details.

We used the synthetic images only on Task 3 (on the two submissions using external data). We added the synthetic images to the training/training splits (never to the holdout or to the training/validation splits) keeping a 1:1 per class proportion (i.e., one synthetic image for each real image in each lesion class). 

% Comment(Fabio)
% dl-11: 4x Tesla P100 (12GB)
% dl-10: 2x Titan Xp (12GB)
% dl-09: 1x Titan Xp (12GB)
% dl-08: 1x Titan Xp (12GB)
% trancoso: 2x Titan Xp (12GB)
% dl-04: 2x GTX Titan X (12GB)
% Azure: 1x Tesla K80 (12GB)
%        1x Tesla P40 (24GB)
\subsection{Computational Resources}
     
To perform a large number of trials, we attempted to secure as much computational horsepower as possible. For deep learning, that means large-memory CUDA-compatible GPUs.

For the experiments, we used NVIDIA GPUs available at RECOD Lab: two Titan X Pascal, six Titan Xp, one Tesla K40, and for Tesla P100. We also used the NC6 (Tesla K80) and ND6 (Tesla P40) virtual machines provided by the Microsoft Azure Cloud platform. 

Although there was a long phase of preliminary experiments, the training and testing of the final models that composed the submissions took only around ten days.

\section{Task 1: Lesion Boundary Segmentation}

This is our second participation in the segmentation task. Although we have some experience in this area, lesion segmentation is not the primary research line of our group. From our previous participation, we decided to keep the  U-shape networks and moderate training times. Contrarily to our previous experiments, showing little  difference between using low (128$\times$128) or high (256$\times$256) resolution, we opted risking for a possible small improvement given by the latter. That was motivated by the knowledge the best networks are tied so closed to the inter-human agreement, and that even small contributions could help. 

Observing the generated masks in preliminary experiments and the 2018 ground-truth annotations --- together with the introduction of a threshold --- we decided for a less fine-grained and more conservative approach concerning details of the final generated mask. To enhance the results, we used a post-processing techniques, to fill the holes in the masks with a morphological operation.

\subsection{Experiments}

We worked on two main models: the FusionNet\footnote{Using this implementation: github.com/GunhoChoi/FusionNet-Pytorch}~\cite{quan2016fusionnet}, a deep fully residual neural network designed for image segmentation in connectomics, and a U-Net-like model\footnote{Code based on github.com/ternaus/TernausNet}~\cite{arXiv:1801.05746}, a convolutional neural network traditionally used for biomedical-image segmentation, with a VGG-16~\cite{simonyan2014very} encoder pretrained on the ImageNet dataset. We trained our models with two datasets: i) Challenge data; and ii) Challenge data plus external data.

Each model was trained using Adam optimizer, using the Cyclic Learning Rate technique~\cite{smith2017cyclical}, on which the learning rate cyclically vary within reasonable boundaries, improving the accuracy and reducing the training time by allowing the model to scape local minima faster. For the cyclic learning rate technique, we used a base learning rate equal to $10^{-5}$ and a maximum learning rate of $10^{-4}$ with a step size of 500. We trained the models for 100 epochs each, with early stopping with patience of 20 epochs. For the loss function, we used the Binary Cross Entropy with soft Jaccard index, with Jaccard weight of 1.0.

We tested four main configurations for the competition: i) FusionNet using only the Challenge data; ii) FusionNet using the Challenge data and external data; iii) U-Net using only the Challenge data; and iv) U-Net using the Challenge data and external data. From our experiments, we noticed that when using external data during the training phase, the results were significantly worse. The large inter-human variability and the existence of several types of ground truths may explain why the task works best on a smaller, but better curated subset of training~data.

After training all the desired networks, we designed the strategy for our submissions, which includes ensembling with our models and post processing the decision. The chosen models were averages, we filled the holes with a morphological operation, and the segmentation mask was upsampled to the image original size.

Our three submissions were (1) average of FusionNet trained on Challenge only, and U-Net trained on Challenge only; (2) average of FusionNet trained on Challenge only, U-Net trained on Challenge only, and FusionNet trained on Challenge and external data; (3) U-Net trained on Challenge only. Our final results on the official testing set were, respectively, 0.694, 0.686 and 0.728 for the threshold Jaccard index. Also, our positions of each submission were, respectively, 88th, 93th and 56th among 112 submissions.

\section{Task 2: Lesion Attribute Detection}

We addressed the task as a patch classification problem rather than a segmentation problem, since our team has a much stronger background in the former, including precoded and pretrained models. 

\subsection{Experiments}

Each image contains about 1,000 superpixels, identified with the same algorithm used in the Challenge to create the ground truths. To address the extreme dataset imbalance, we train 750 balanced batches per epoch, for 30 epochs on two different internal splits for each model. 

First, we crop the images into patches, each with a superpixel in its center. The patch dimensions are one of the factors evaluated: 128$\times$128 and 299$\times$299. By employing bigger patches, we expect the network to learn not only from the center superpixel, but also from its neighborhood. 

We fine-tune an Inception-v4~\cite{szegedy2016inceptionv4} network pretrained on ImageNet. We employ Stochastic Gradient Descent with a momentum of 0.9, batches of size 16, and starting learning rate of 0.001, decreasing it to 0.0001 after epoch 12. Data augmentation is one of the factors we tested. When applying augmentations on training (random flips, rotations, and color jitter) the result was significantly worse. We suspect that these augmentations could displace the superpixel from the center of the patch. Our final submission does not contain any data augmentation during training. We keep the augmentation for test with 16 replicas at all models, with random flips and color~jitter.

The network learns to classify each of the crops (which are linked to a superpixel of the image) as one of the six classes: absent, pigment network, negative network, streaks, milia-like cyst, and globules. To generate the predictions, the test set also needs to be cropped into patches. Each patch receives a prediction about its class. We classify the patch by selecting the class with the highest score assigned by the network.

Next, we compose the masks from the predictions and apply a post-processing procedure to eliminate positive superpixels given a threshold (30 in our experiments showed the best result). This is used to attenuate false positives that occurred especially in the most abundant class, absent.

Our three submissions were (1) average of 4 best deep learning model with final thresholding; (2) average of 4 deep learning models, without final thresholding; (3) the single best model on the training/validation split with post-processing. Our final results on the official testing set were, respectively, 0.344, 0.337 and 0.323 for the Jaccard index. Also, our positions of each submission were, respectively, 14th, 15th and 17th among 26 submissions.

Although the patch classification approach allowed our team to participate in this task, it is \textit{very} time-consuming: testing takes longer than training! As a consequence, we separated the holdout set but did not have time to evaluate the models on it. We employed the less-than-ideal training/validation performance to select the models.

\section{Task 3: Lesion Diagnosis}

Automated lesion classification is the most traditional research line in our group. This year, our explorations started from our participation on ISIC Challenge 2017~\cite{menegola2017recod} and our follow-up research~\cite{valle2017data}. We also look for novelties and insights from the Machine Learning community that could bring new competitive gains. Although many of the experiments were performed systematically, simulating a factorial design, not all combinations were evaluated. Also, the training of some models were limited due to time and computational resources.

\subsection{Experiments}

We trained three different CNN architectures: Inception-v4~\cite{szegedy2016inceptionv4}, ResNet-152~\cite{he2016deep}, and DenseNet-161~\cite{huang2017densely}, all pretrained on ImageNet dataset. We fine-tuned the networks on three datasets: full, only, and full+synthetic augmentation~\cite{bissoto2018skin}.

Each network was trained with Stochastic Gradient Descent with momentum of 0.9, batch size of 32, starting learning rate of $10^{-3}$, multiplied by 0.1 whenever the validation loss fails to improve for 10 epochs, until it reaches $10^{-5}$. Images were resized online to 224$\times$224 for ResNet and DenseNet, and to 299$\times$299 for Inception-v4. We normalized the images by subtracting the mean and dividing by the standard deviation channel-wise.

To deal with dataset imbalance, we set the optimization goal to a class-weighted cross-entropy, with the weights calculated by dividing the frequency of the most common class by the frequency of each class. We applied early stopping with a patience of 22 epochs by monitoring the validation loss. 

We performed online data augmentation as described in~\cite{perez2018data} (scenario J): random crops (preserving 0.4-1.0 of the original area, and 3/4-4/3 of the original aspect ratio); random vertical/horizontal flips; rotation (0-90°); shear (0-20°); area scaling (0.8-1.2); random color transformations on saturation, brightness, contrast, and hue. We applied the transformations to the validation (single replica), holdout (32 replicas), and final test (128 replicas), taking the decision as the average of the replicas.

Our three submissions were (1) XGBoost ensemble of 43 deep learning models; (2) average of 8 best deep learning models (on the holdout set) augmented with synthetic images\footnote{N.B. that approach is wrongly named as an average of \textit{15 models} on the official leaderboard} and (3) average of 15 deep learning models trained only with Challenge data. Our final results on the official testing set were, respectively, 0.732, 0.725 and 0.803 for the normalized multi-class accuracy. Also, our positions of each submission were, respectively, 32th, 39th and 9th among 141 submissions. 

% FECHAMOS EM:
%@fabioperez Acho que poderíamos mandar uma só com o ensemble. Estou achando cada vez mais mico. Mesmo que ele saia, a chance de não ficar legal é grande. Sugestão:
%(1) Ensemble de todas (já que média de todas não é uma boa ideia)
%(2) Média do Alceu (filosofia DDD)
%(3) Média das only (para concorrer na caregoria only)

% COMENTAR QUE: 
% PAra curiosidade: simplesmente aplicar a camada de XGB (sem ensemble) nos modelos do Fabio, piorou ligeiramente os resultados. Para mim isso reforça os resultados do nosso paper Data Depth Design, de que colocar camada final em Deep é mico.

\section{Final Comments}

We are very excited to see the ISIC Challenge as a continuing event, since we consider such initiative as pivotal for the development of our research area. Until recently, making comparisons across different approaches for skin lesion analysis was essentially impossible, due to difficulties of code and data sharing, and lack of standardized evaluation metrics and datasets~\cite{fornaciali2016towards}. We also acknowledge the importance of keeping the testing set secret until all evaluations were over, preventing, thus, subtle methodological errors that inflate the performance evaluation of models~\cite{valle2017data, salzberg1997comparing}. 

Despite the diversity of skin lesion types and their dermatological importance, we asked ourselves whether making the classification task (Task 3) so fine-grained was really necessary, especially given the huge class imbalance. We hope to be surprised when the results become public, but we fear that confusion among very small classes (e.g., Benign Keratosis and Actinic Keratosis) will bring much noise to the evaluation. In our current research, we are still focusing on coarse-grained melanoma/non-melanoma screening/triage classifiers --- and we notice that real-world performances even for such coarse-grained procedures are still far from ideal. 

We noticed the variability of the annotations as an important difficulty for Task~1. While some lesions were very finely annotated, others are merely polygons around the lesion. The performances on that task are --- or at least were, in 2017 --- close to the limit of inter-human agreement, and those different ``definitions'' of what is a segmentation bring extra fluctuations. As a suggestion, maybe using the convex hulls of the human annotations is sufficient for location purposes, and provides a less noisy target for comparing algorithms.

Despite our lack of experience, we were excited to participate in Task 2. It is a new problem for the automated skin lesion analysis community and poses several challenges: especially in terms of evaluating the models, given the hugely unbalanced annotations. For us, however, the existence of that task had an additional importance: its ground-truth annotations allowed us to create, for the first time, realistic synthetic lesion images, with proper dermatologic configurations, using Generative Adversarial Networks~\cite{bissoto2018skin}. As this opens a new frontier in dealing with the scarcity of annotated data, we hope the community will work to provide more of this valuable type of ground truth.

\section*{Acknowledgements}

{\small
A. Bissoto is funded by CNPq; M. Fornaciali and E. Valle are partially funded by Google Research Awards for Latin America 2017; E. Valle is also partially funded by a CNPq PQ-2 grant (311905/2017-0) and Universal grant (424958/2016-3). RECOD Lab. is partially supported by diverse projects and grants from FAPESP, CNPq, and CAPES. We gratefully acknowledge NVIDIA Corporation for the donation of GPUs and Microsoft Azure for the GPU-powered cloud platform used in this work. We are grateful to RECOD members --- and in particular to Pedro Tabacof and Ramon Oliveira --- for scientific and technical insights on machine learning. We thank Prof. Flávia V. Bittencourt and Prof. Gabriela Salvio for kindly providing additional skin lesion images. We also thank our partners from the School of Medical Sciences of UNICAMP for the dermoscopic discussions that enhanced our comprehension about skin lesion images. }
   
%\balance
%\vspace{-0.1cm}	
\bibliographystyle{IEEEtranN}
\bibliography{references}
    
\end{document}